\documentclass[10pt,twocolumn,letterpaper]{article}

\usepackage[normalem]{ulem}
\usepackage[pagenumbers]{cvpr} 

\usepackage{overpic}
\usepackage{tikz}

\usepackage{xcolor}

\newcommand{\goldmedal}{\raisebox{-0.2ex}{\includegraphics[height=2.0ex]{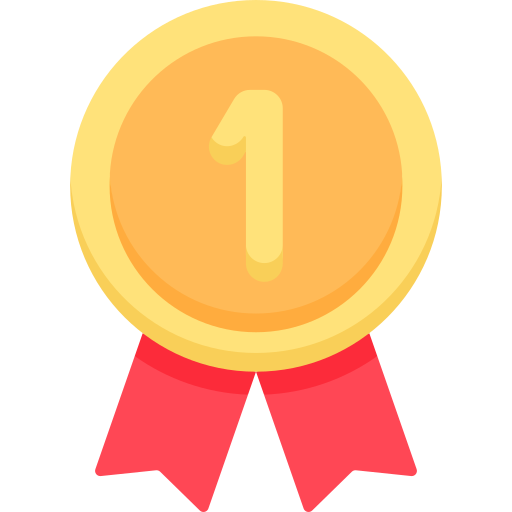}}}
\newcommand{\silvermedal}{\raisebox{-0.2ex}{\includegraphics[height=2.0ex]{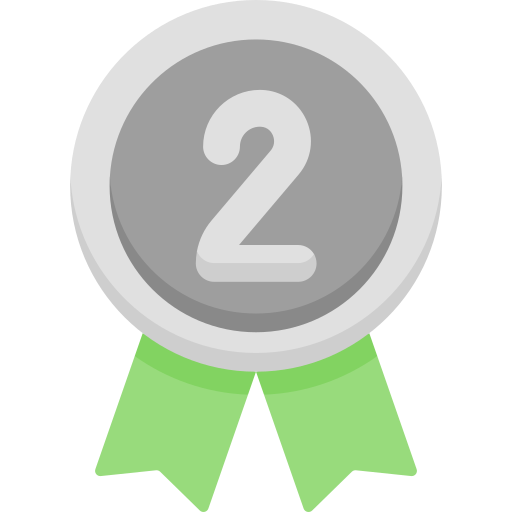}}}
\newcommand{\bronzemedal}{\raisebox{-0.2ex}{\includegraphics[height=2.0ex]{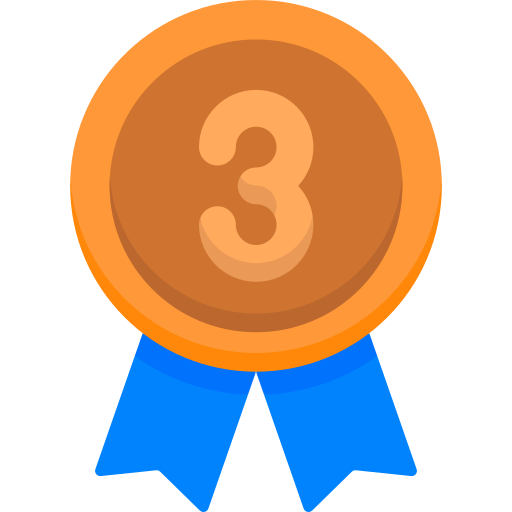}}}

\definecolor{cvprblue}{rgb}{0.21,0.49,0.74}
\usepackage[pagebackref,breaklinks,colorlinks,allcolors=cvprblue]{hyperref}

\def\paperID{*****} 
\def\confName{CVPR}
\def\confYear{2026}

\title{Edge Prediction for Roof Wireframe Reconstruction with Transformers}

\author{Gustav Hanning${}^*$, Ludvig Dillén${}^*$, Jonathan Astermark${}^*$, Johanna Lidholm${}^*$, Viktor Larsson\\
Centre for Mathematical Sciences, Lund University\\
{\tt\small \{gustav.hanning,ludvig.dillen,jonathan.astermark,johanna.lidholm,viktor.larsson\}@math.lth.se}
}

\begin{document}
\maketitle
\begin{abstract}
This paper presents a competitive solution to the S23DR Challenge 2026, which aims to reconstruct 3D house roof wireframe models from sparse SfM point clouds and ground-level semantic segmentations and depth maps. Our proposed method utilizes an end-to-end Transformer encoder-decoder architecture inspired by DETR. To effectively process the geometric and semantic data, the sparse SfM point cloud input is dynamically subsampled based on semantic priority and augmented with Gestalt and ADE20k class features. 
To further increase segmentation context, we fuse the point features with additional Gestalt feature encodings
which are obtained by projecting the points into latent feature maps produced by a frozen autoencoder.
Learned query embeddings are then decoded directly into 3D wireframe edges via cross-attention mechanisms.
Evaluated on the "HoHo 22k" dataset, our approach significantly outperforms both handcrafted and learned baselines, achieving a Hybrid Structure Score (HSS) of 0.6476 and securing the second-highest position on the challenge's private leaderboard.
\end{abstract}    
\let\thefootnote\relax\footnote{* Equal contribution}
\section{Introduction}
\label{sec:intro}

We present our solution to the Structured Semantic 3D Reconstruction (S23DR) Challenge 2026 \cite{s23dr2026}. The goal of the challenge is to reconstruct house roof wireframe models given sparse Structure-from-Motion (SfM) point clouds and ground-level Gestalt and ADE20k~\cite{jain2023oneformer,zhou2019semantic} semantic segmentation images and depth maps.

Our method is based on a simple Transformer \cite{vaswani2017attention} architecture inspired by DETR \cite{carion2020end}. We train the network end-to-end to directly predict wireframe edges from the input SfM point cloud, which is augmented with features from the segmentation images and subsampled to include the most relevant points.
The point cloud is passed through our encoder-decoder model, where a set of learned queries are decoded into edges.
We additionally fuse the latent point features with multi-view informed Gestalt feature vectors, encoding local segmentation details, which are obtained by projecting the points into feature maps  
generated by a frozen autoencoder.

\begin{figure*}[t]
    \centering    
    \begin{overpic}[width=\linewidth]{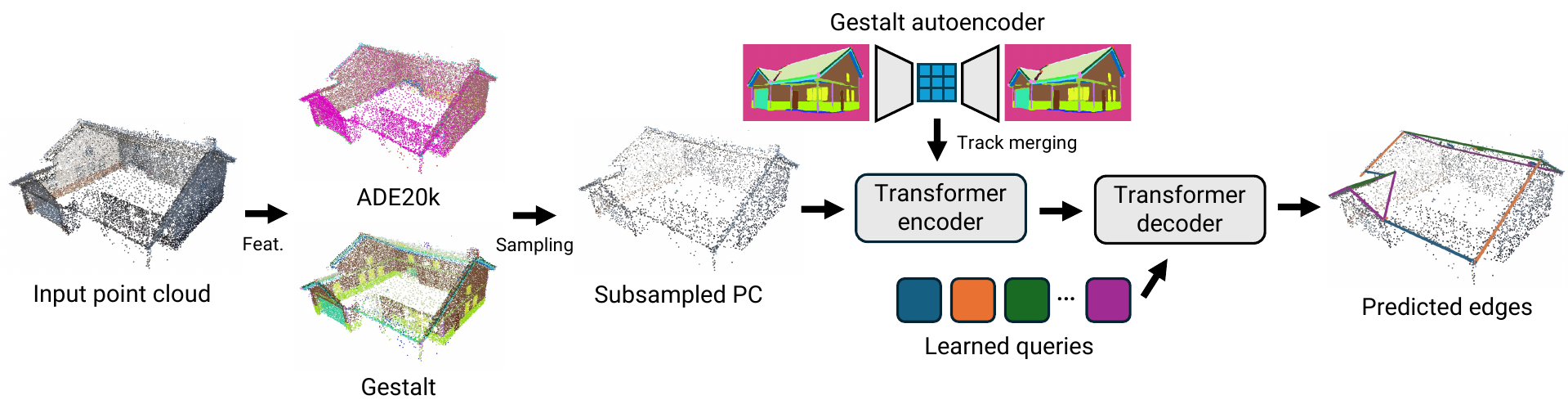}
    \end{overpic}
    \caption{Our method predicts rooftop wireframe edges from a point cloud using a transformer encoder-decoder network.
    }
    \label{fig:overview}
\end{figure*}

The solution achieves the second-highest score on the private leaderboard. An overview of our method is shown in \cref{fig:overview}.

\section{Dataset}
\label{sec:dataset}
In this section, we describe the "HoHo 22k" dataset and the evaluation protocol used in the challenge. We also summarize the main differences compared to last year's challenge and discuss some data-related issues encountered.
\begin{figure}[t]
    \centering    
    \begin{tikzpicture}[scale=1.0]
        \node at (2.7,0.0) {\includegraphics[scale=0.2]{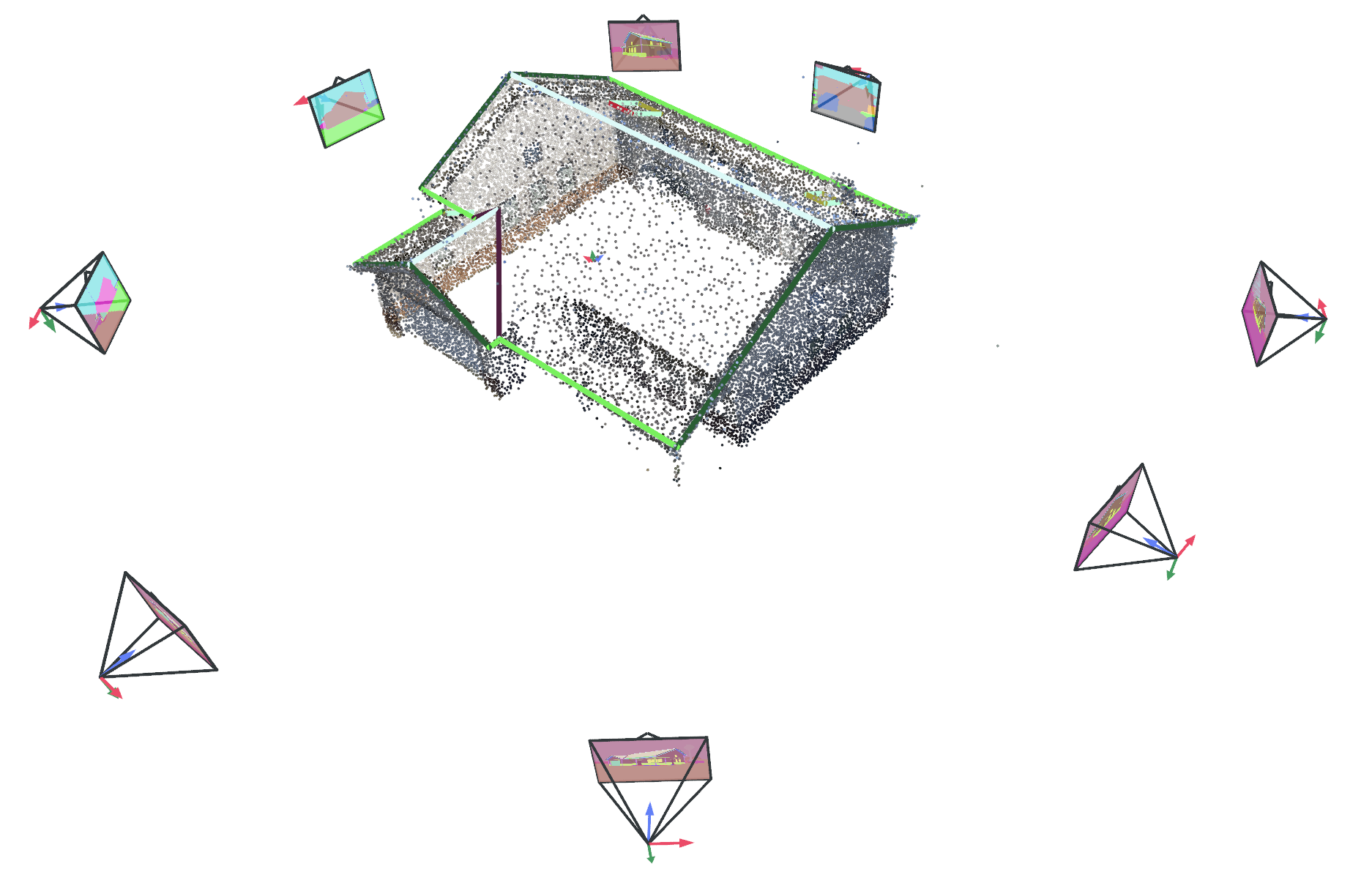}};
        \node at (0.0,-3.3) {\includegraphics[scale=0.1]{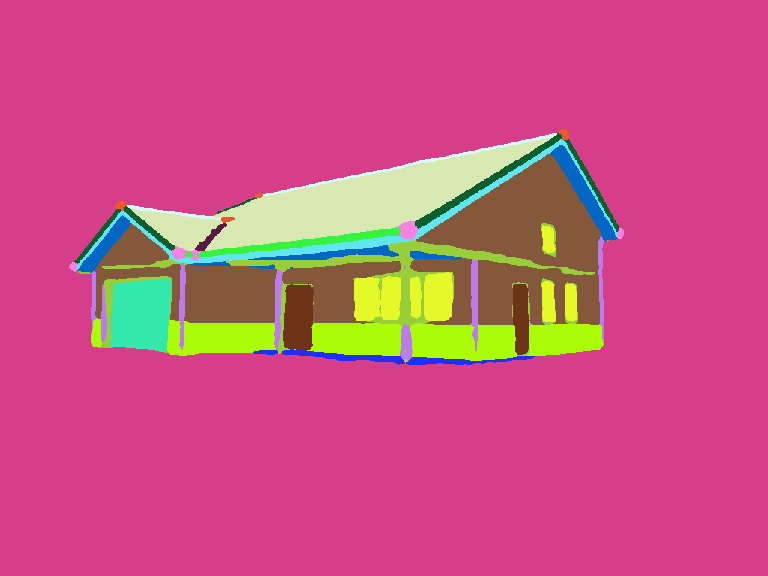}};
        \node at (2.7,-3.3) {\includegraphics[scale=0.1]{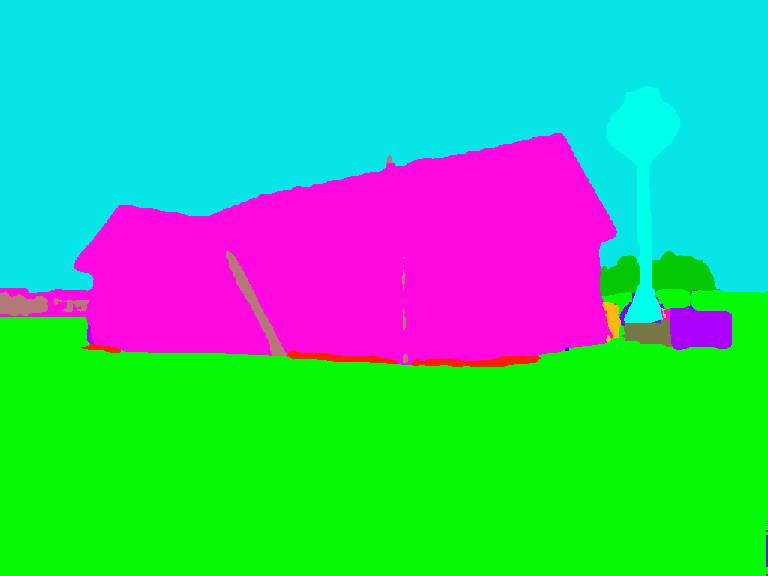}};
        \node at (5.4,-3.3) {\includegraphics[scale=0.1]{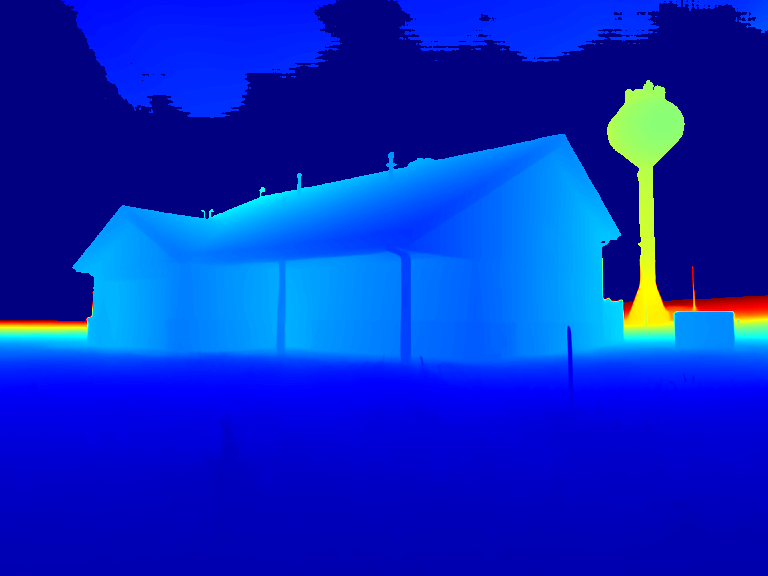}};
    \end{tikzpicture}
    \caption{Example scene from the HoHo 22k dataset. \textbf{Top}: Sparse SfM point cloud and cameras. \textbf{Bottom}: Gestalt/ADE20k segmentations and depth map for one of the views.}
    \label{fig:hoho2k}
\end{figure}

\subsection{Modalities}

Each scene in the provided dataset consists of the following:

\begin{itemize}
    \item A 3D point cloud generated by a VGGT-based \cite{wang2025vggt} SfM pipeline, with tracks, intrinsics and poses.
    \item Gestalt and ADE20k segmentations for the images.
    \item MoGe-2 \cite{wang2026moge} depth maps.
    \item Ground truth wireframes (vertices and edges) and semantic edge classes.
\end{itemize}
The original images are not included in the dataset. An example scene is shown in \cref{fig:hoho2k}.

\subsection{Splits and Submissions}

The dataset is split into a training set with 19677 scenes, a validation set with 170 scenes, and a test set only accessible through the Hugging Face platform. The test set consists of a public split on which participants are allowed 5 submission attempts per day (each with a 2-hour runtime limit) and a hidden split announced at the end of the challenge. Each team is allowed to submit two of their solutions for evaluation on the hidden test split. The performance on the hidden split determines the winner.

\subsection{Metrics}

The evaluation metric used is called Hybrid Structure Score (HSS)~\cite{langerman2025explaining}, which is the harmonic mean of the F1 score of the vertices and the intersection-over-union (IoU) of the edges. A vertex is predicted positive if it is within 0.5 m from a ground-truth vertex. The edges are modeled with a 0.5 m radius cylinder for deciding the IoU score.

\subsection{Updates from S23DR 2025}

The dataset went through a major cleanup for this year's challenge. The original dataset was reviewed and inconsistent scenes were removed, resulting in a total of 22k scenes compared to last year's 25k. The pose ambiguity reported by participants from last year~\cite{skvrna2025structured} seems to have been resolved, as only one unique set of camera parameters is provided per camera. The reconstruction quality was improved and more images per scene included. Last year's depth maps were generated from Metric3D \cite{hu2024metric3d} but are now produced by MoGe-2 \cite{wang2026moge}. The images were reported to be downscaled to 768 px for compute and storage reasons.

\subsection{Data-related Challenges}
Working with the challenge, we have encountered a few dataset issues. This includes problems with downloading the dataset as well as the quality of the data.

\textbf{Problem downloading the dataset:}
Due to a few corrupt images in the training set, we were first only able to download around 13k scenes with the Hugging Face dataset library. After implementing a workaround in our code, we could eventually train our model on the full set, minus the corrupt samples.

\begin{figure}[t]
    \centering    
    \includegraphics[width=\linewidth]{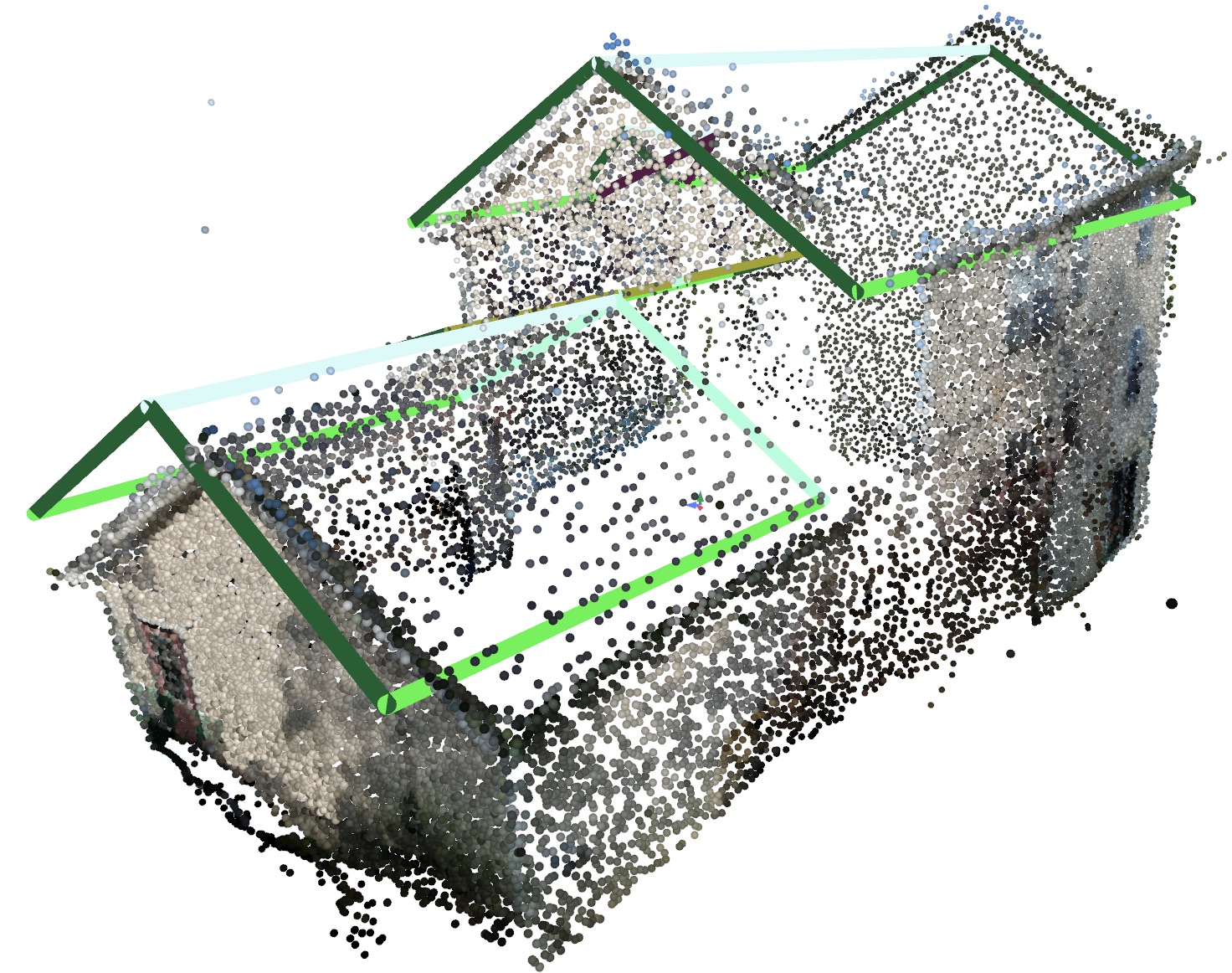}
    \caption{The ground truth wireframe is misaligned with the SfM point cloud. This example is from the validation set.}
    \label{fig:misaligned}
\end{figure}

\textbf{Ground-truth wireframe  misalignment:}
We observed that some scenes in both the train and validation splits had misalignment issues between the point cloud and the ground-truth wireframe; see~\cref{fig:misaligned} for an example.
For this reason, we manually went through the validation dataset and removed 32 misaligned scenes. For the training data, a manual approach was not feasible so we took two different approaches. First we excluded the samples identified by the organizer's learned baseline \cite{langerman2026s23drbaseline}. Then we ran our current best model at the time on the training set and filtered away all scenes with HSS $<0.01$. That gave a reduction of training scenes by 781. The filtering was conservative as we did not want to remove hard training scenes.

\section{Method}
\label{sec:method}

Our method takes as input a point cloud augmented with Gestalt and ADE20k class features, subsampled to a fixed number of points (\cref{subsec:pre-processing}). The network architecture follows a standard encoder-decoder design with a set of learned queries (\cref{subsec:network-architecture}).
To increase the receptive field into the segmentation images, we fuse the point features with multi-view informed Gestalt feature vectors, encoding a richer segmentation context (\cref{subsec:gestalt-fusion}).
We train the network (\cref{subsec:training-details}) with cross entropy and $L_1$ loss (\cref{subsec:loss-function}). The predicted edges are post-processed to form the final wireframe model (\cref{subsec:post-processing}).

\subsection{Pre-processing}
\label{subsec:pre-processing}

First, each point in the sparse SfM point cloud is projected into every view in its track. The Gestalt and ADE20k classes are determined by majority vote and are one-hot encoded. 
We then subsample the point cloud to a fixed number of points, $N_p$. Points belonging to the Gestalt edge and vertex classes, which are the most informative for the task, are sampled with probabilities 10 and 100 times higher, respectively, than those of other points.
Conversely, points with Gestalt class "unclassified" or "unknown" are assigned zero probability. The result is a subsampled point cloud $P \in \mathbb{R}^{N_p \times 3}$ with features $F \in \mathbb{R}^{N_p \times D}$, where $D$ is the point feature dimension. The one-hot encoded Gestalt and ADE20k features are of dimension 28 and 149, each, and we also include the RGB color (normalized to the $[0, 1]$ range), so $D = 28 + 149 + 3 = 180$. Finally, the point cloud is resized to fit in the unit cube.

\subsection{Network Architecture}
\label{subsec:network-architecture}

Our network has a standard encoder-decoder structure similar to DETR \cite{carion2020end} but adapted to the problem of roof wireframe reconstruction. We project the input features $F$ into an embedding space of dimension $D_e$, add fixed positional encodings \cite{tancik2020fourier} computed from to the point cloud $P$ and pass the projected features to a Transformer encoder.

The encoded point cloud is, together with a set of learned anchor edges and corresponding query embeddings, the input to a Transformer decoder. Self-attention among edges and cross-attention between edges and points are used alternately in the decoder layers. Positional encoding is applied for both points and edges at each step.

There are two prediction heads in the network: a classification head and a coordinate head. The classification head is a simple linear layer that outputs logits for the edge classes. As the metrics used in the challenge are purely geometrical, we do not try to predict the actual edge class but instead utilize a single class plus a "background" class. The coordinate head is a small MLP followed by sigmoid activation, predicting the normalized coordinates of the two endpoints.

\subsection{Local Gestalt Feature Fusion}
\label{subsec:gestalt-fusion}

While each point is assigned a Gestalt and ADE20k class by majority voting in the pre-processing step, this will inevitably not capture all of the information contained in the segmentation masks due to the sparsity of the point cloud.
This should be especially true for the Gestalt maps, which contain explicit segmentations of edges and vertices.
In order to increase the receptive field of the points, and recover more of the Gestalt segmentation context, we fuse each projected point feature with an additional learned Gestalt feature that is both multi-view informed and encodes a richer local context.

First, each Gestalt image is encoded to a coarse feature map using a frozen autoencoder with a 32$\times$32$\times$32 bottleneck (\ie, a spatial 32$\times$32 grid of patches, each with feature dimension 32).
After subsampling the point cloud, we retrieve a set of corresponding 32-dimensional features based on the projections into each view.
Each feature is then projected to a $D_g$-dimensional vector using a shared linear projector. A positional encoding of the relative view vector, defined as the normalized direction from the origin (in resized point space) to the camera center, is added to this $D_g$-dim vector.
The feature vector is averaged over all views in the track, and then passed through a final linear layer to get the final $D_e$-dim Gestalt feature.
This Gestalt feature is fused with the projected point feature before the encoder by addition, and weighted with a learned factor which is initialized to 0.
For simplicity, we used $D_g = D_e$ and the same positional encoder as for point positions.

\subsection{Loss Function}
\label{subsec:loss-function}

We supervise directly on the predicted wireframe edges. They are matched to the ground-truth edges using the Hungarian algorithm \cite{kuhn1955hungarian} with the matching cost between a ground truth edge $e$ and prediction $\hat e$ given by
\begin{equation}
    \mathcal{C}(e, \hat e) = -\hat p_c + \mu \min \left( \| e - \hat e \|_1, \| \tilde e - \hat e \|_1 \right),
\end{equation}
where $\hat p_c$ is the predicted probability that $\hat e$ belongs to the class $c$ of $e$, and $\mu$ is a scale factor. We compute the $L_1$ distance over the six coordinates, corresponding to the two endpoints of the edge. Since edges are undirected, we take the minimum distance over the original ground-truth edge $e$ and its flipped version $\tilde e$,
\ie the edge $e$ with its endpoints reversed.

Assuming that the number of queries $N_q$ is larger than the number of ground truth edges $N_g$ the result of the Hungarian matching is a set of edge correspondences $\{ (e_i, \hat e_i) \}_{i=1}^{N_g}$ and unmatched predictions $\{ \hat e_i \}_{i=N_g + 1}^{N_q}$. From these we calculate a loss $\mathcal{L} = \mathcal{L}_{CE} + \lambda \mathcal{L}_{L_1}$, where
\begin{equation}
    \mathcal{L}_{L_1} = \frac{1}{N_g} \sum_{i=1}^{N_g}  \min \left( \| e_i - \hat e_i \|_1, \| \tilde e_i - \hat e_i \|_1 \right)
\end{equation}
is the average $L_1$ distance for matched edges and
\begin{equation}
    \mathcal{L}_{CE} = - \frac{1}{\sum_{i=1}^{N_q} w_{c_i}} \sum_{i=1}^{N_q} w_{c_i} \log \hat p_{c_i}
\end{equation}
a weighted cross-entropy loss. Here, $w_{c_i}$ is the weight for the class $c_i$ of the matched ground truth edge $e_i$ or the background class if there was no match.

\subsection{Implementation and Training Details}
\label{subsec:training-details}

We train our network on our cleaned HoHo 22k training set. During training, two types of data augmentations are applied to prevent overfitting: random rotation around the $y$-axis (up direction) and a 50\% probability of mirroring the point cloud in the $yz$-plane. The weighted sampling described in \cref{subsec:pre-processing} also acts as a regularizer and $N_p = 7168$ points are sampled for each scene to create training examples. We use $N_q = 100$ queries and set $\mu = \lambda = 5$ for matching and computing the loss $\mathcal{L}$, which is applied to the output of each decoder layer and summed. The class weights are $w_e = 1$ and $w_b = 0.1$ for the edge and background class, respectively.

The network has 5 encoder and 5 decoder layers, with embedding dimension $D_e = 360$. It is trained for 75 epochs with the AdamW \cite{kingma2014adam,loshchilov2017decoupled} optimizer using a weight decay of $10^{-4}$ and a one-cycle learning rate schedule \cite{smith2019super}. The learning rate is initially set to $2 \times 10^{-5}$, anneals to a maximum of $2 \times 10^{-4}$ and then gradually decreases to the final value $2 \times 10^{-7}$. The batch size is 11. Dropout with probability 0.1 is applied in both encoder and decoder layers, along with layer normalization \cite{ba2016layer}. Every 500 iterations we compute the mean HSS on our cleaned validation set and save the weights that maximize this metric to use for inference. The model takes 12 hours to train on an RTX 4090 GPU and has 23M parameters.

\textbf{Gestalt autoencoder:}
The encoder part of the autoencoder consists of four layers, each applying a $4 \times 4$ convolution with stride 2, followed by batch norm \cite{ioffe2015batch} and a ReLU activation.
The number of output channels are $\{32, 64, 128, 256\}$, respectively.
A final $1\times1$ convolution projects to the 32-channel bottleneck.
The decoder architecture is the transpose of the encoder, but outputs class logits which are then are mapped back to the RGB values of the winning class.

For training the autoencoder, 1000 scenes are randomly selected from the train set, giving a total of 9422 training images.
Each image is reshaped to 512$\times$512, giving the bottleneck size of 32$\times$32 with a receptive field of $46\times46$ pixels.
Training is done using weighted cross-entropy loss, where edge and vertex classes are assigned a weight of 10 and 100, respectively, while the rest of the weights are set to 1.
We use the same optimizer and learning rate schedule as for the Transformer training, and train for 50 epochs with batch size 24.

\subsection{Post-processing}
\label{subsec:post-processing}

At inference time, edges predicted to be of the background class and those with probability $\hat p_c$ less than 0.95 are first removed. The result is a set of disconnected edges from which we construct the house roof wireframe.

Since each predicted edge is represented by two independent endpoints, multiple endpoints may be predicted near the same physical roof corner. To avoid duplicate vertices, we merge nearby predicted vertices after edge prediction using an iterative centroid-based strategy: (1) calculate the distances between all vertices; (2) update the two closest vertices to their average vertex position, given that they are closer than 0.5 meters, keeping only one of the vertices.
The algorithm repeats until no pair of vertices are closer than the distance threshold.
Note that we do not allow the two endpoints of the same predicted edge to merge, since that would collapse an edge into a point. Duplicate edges may appear following this procedure, and all but one copy are removed for each edge.

Our model uses absolute positional encoding, making it sensitive to rotations and flips, even though the underlying reconstruction task should be invariant to such transformations. Since inference is relatively fast, taking only about 25 minutes of the allowed 2 hours on the evaluation server, we employ test-time augmentation (TTA) to improve robustness. We also ensemble two independently trained checkpoints of the same model. In total, we evaluate both checkpoints over four yaw rotations, \(0^\circ, 120^\circ, 240^\circ,\) and \(300^\circ\), with and without horizontal flipping, yielding 14 augmented predictions (only two variants for $300^\circ$ due to resource constraints).

\section{Results}
\label{sec:results}

In this section, we present quantitative and qualitative results of our wireframe prediction model. We present the baselines provided, results on the public and private test sets and some ablation studies.

\subsection{Baselines}

Two baselines are given by the workshop organizers: one handcrafted \cite{usm3d2026handcrafted} and one learned \cite{langerman2026s23drbaseline}. In short, the handcrafted baseline detects vertices and edges in the Gestalt segmentations and lifts these 2D detections to 3D using depth maps that were refined to fit the SfM point cloud. As the same vertex/edge can typically be seen in multiple views, the lifting is followed by a de-duplication step.

The learned baseline utilizes a Perceiver-based \cite{jaegle2021perceiver} transformer architecture trained on fused 3D point clouds, created by combining the SfM point cloud with unprojected depth maps. Similar to our approach, it uses Gestalt and ADE20k class features, but not RGB color. The network is trained in a manner comparable to ours, matching predicted and ground-truth edges and using coordinate and cross-entropy loss, but in contrast employs a more complex three-stage training procedure.

\subsection{Challenge Results}

\begin{figure}[t]
    \centering    
    \begin{tikzpicture}[scale=1.0]
        \node at (0.0,0.0) {\includegraphics[scale=0.063]{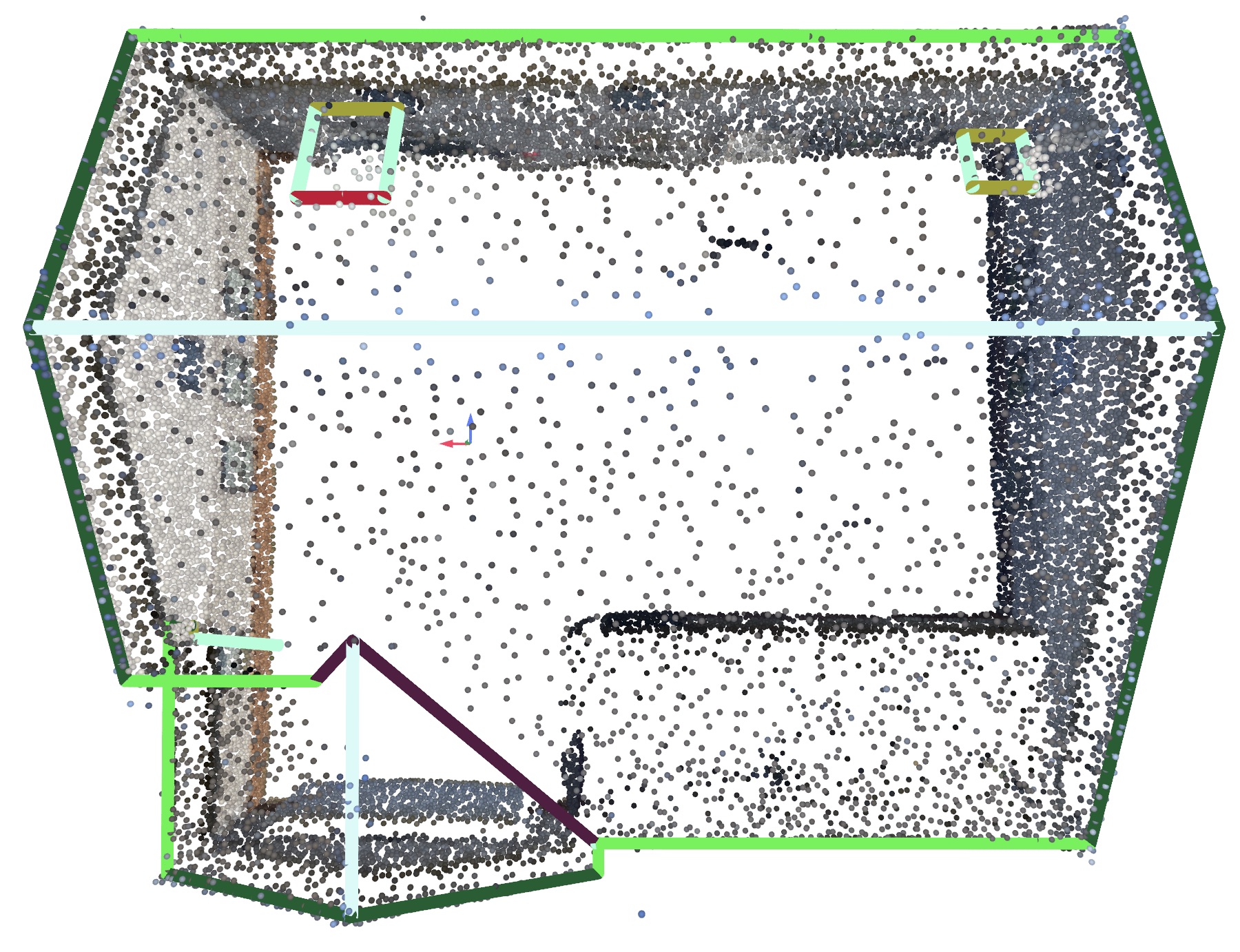}};
        \node at (4.1,0.0) {\includegraphics[scale=0.063]{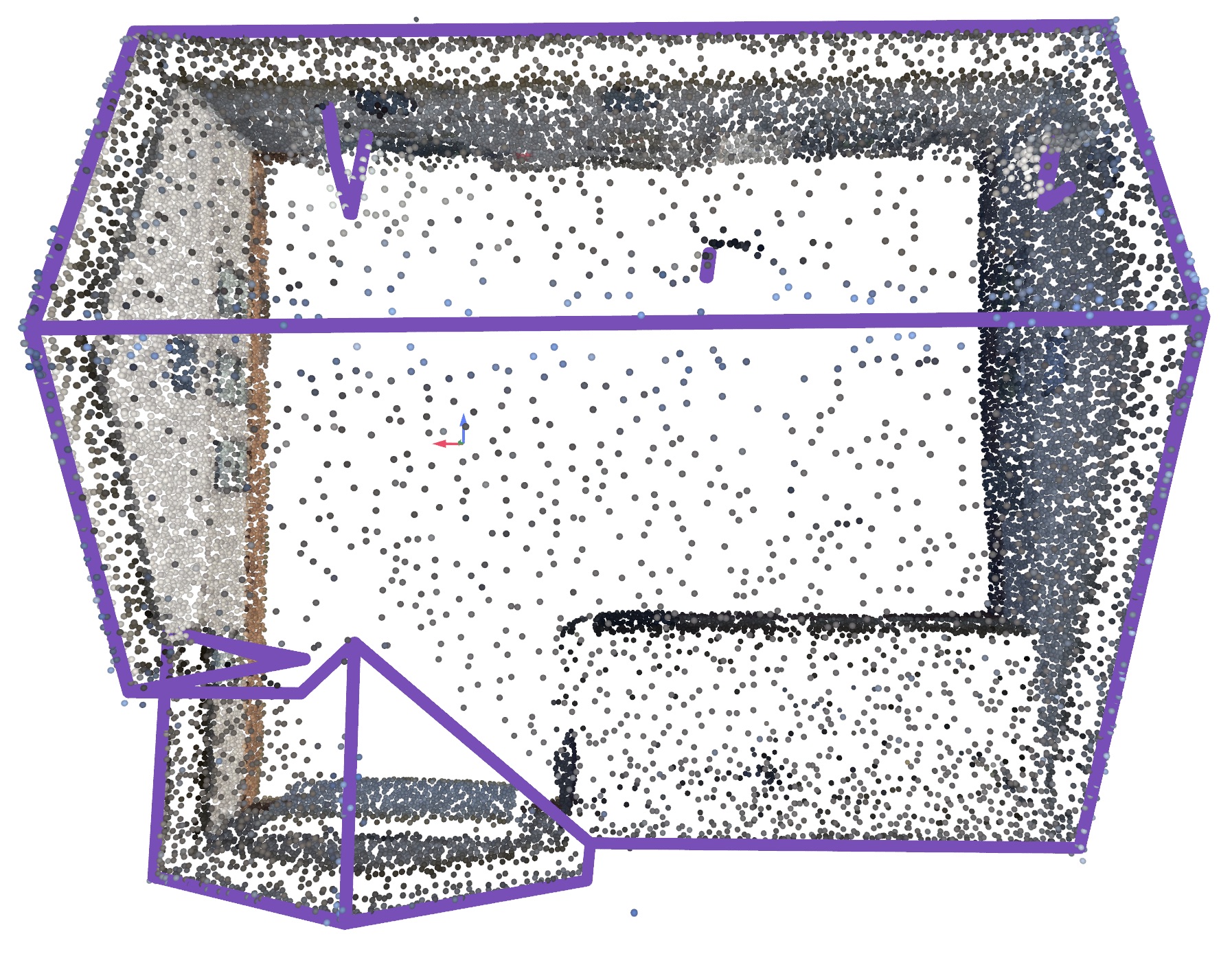}};
    \end{tikzpicture}
    \caption{Ground truth (\textbf{left}) and predicted (\textbf{right}) roof wireframes for one scene in the HoHo 22k validation set. Our method can accurately predict longer edges but struggle with smaller details like chimneys.}
    \label{fig:example-pred}
\end{figure}

In \cref{tab:results} we report results on the public and private test sets. Our method significantly surpasses the two baselines, both in terms of vertex F1 score and edge IoU. It is outperformed by the challenge winner ("VRG\_jskvrna") by 0.0066 HSS on the private test set, but is substantially better than the third-place contestant ("StarAtNyte1"). \cref{fig:example-pred} shows example predictions for one of the scenes in the validation set. While our model accurately reconstructs the most significant edges of the roof, it fails to predict some of the shorter edges, for example those around the chimney in the top right part of the two images.

\begin{table*}[t]
  \caption{Results on the public and private test sets.}
  \label{tab:results}
  \centering
  \begin{tabular}{l c c c l c c c}
    \toprule
    & \multicolumn{3}{c}{Public} && \multicolumn{3}{c}{Private} \\ \cmidrule{2-4} \cmidrule{6-8}
    Method & HSS & F1 & IoU && HSS & F1 & IoU \\
    \midrule
    \goldmedal~ \textsc{VRG\_jskvrna} $\quad\quad\quad$ & \textbf{0.6068} & \textbf{0.7351} & 0.5252 &$\quad$& \textbf{0.6542} & \textbf{0.7907} & 0.5665 \\
    \silvermedal~ \textsc{LundUni} (Ours) & 0.6067 & 0.7113 & \textbf{0.5375} && 0.6476 & 0.7506 & \textbf{0.5779} \\
    \bronzemedal~ \textsc{StarAtNyte1} & 0.5746 & 0.6639 & 0.5156 && 0.6060 & 0.6955 & 0.5451 \\
    Learned baseline & 0.4470 & 0.4910 &0.4229 && 0.4739 & 0.5122 & 0.4536 \\
    Handcrafted baseline & 0.3559 &0.4717 & 0.2971 && 0.3907 & 0.5067 & 0.3281 \\
    \bottomrule
  \end{tabular}
\end{table*}

\subsection{Ablation Experiments}

We validate some of our design decision by conducting ablation experiments on the full validation set (\cref{tab:ablations}), where individual components are removed or modified relative to our full model (top row).

First, we run our method with a single trained checkpoint (second row) and compare with the ensemble of two. Using the ensemble is marginally better across all metrics.
Next, the test-time augmentations are also disabled (third row), resulting in a larger performance hit.
A single network that does not use the Gestalt autoencoder is evaluated (fourth row). It is slightly worse than one that includes the Gestalt feature fusion (second row).
We next perform the point sampling with equal probability for all points (except "unclassified" and "unknown", which are still excluded). As we sample a relatively large number of points ($N_p = 7168$), the resulting point clouds still contain many points with Gestalt vertex and edge classification, and the performance decrease is minor (fifth row compared to first). 
Finally, we try running with only a quarter of the points ($N_p = 1792$, now with the higher weights for vertex and edge points), and note that our method is fairly robust to the input point cloud density (last row).

\begin{table}[t]
  \caption{Ablation experiments on the full validation set.}
  \label{tab:ablations}
  \centering
  \begin{tabular}{l c c c}
    \toprule
    Method & HSS & F1 & IoU \\
    \midrule
    Full model & \textbf{0.4845} & \textbf{0.5655} & \textbf{0.4377} \\
    $-$ Ensemble & 0.4823 & 0.5615 & 0.4356 \\
    \quad $-$ Test-time aug. & 0.4690 & 0.5502 & 0.4233 \\
    \quad $-$ Gestalt feat. fusion & 0.4764 & 0.5535 & 0.4335 \\
    $-$ Weighted sampling & 0.4802 & 0.5630 & 0.4330 \\
    $-$ Num. points & 0.4791 & 0.5602 & 0.4322 \\
    \bottomrule
  \end{tabular}
\end{table}








\section{Unsuccessful Approaches}
\label{sec:unsuccessful-approaches}

We describe in this section some of the ideas that we tried, which did not improve the performance of our method.

\subsection{Plücker Lines}
One geometric primitive we tried as network input beyond points was Plücker lines \cite{hartley2003multiple}. The main reason for using Plücker lines was to be able to provide 3D information to the network in regions where the point cloud was sparse, often due to low co-visibility. To obtain Plücker lines, we sampled them from the Gestalt segmentation mask only at predicted edge and vertex classes, with a 10 times higher weight for sampling vertices, as there are significantly fewer vertex pixels than edge pixels. Explicitly, at a pixel sample $x$ in an image, we form the ray from the camera center, $C=-R^Tt$, to $x$ in world coordinates as
\begin{equation}
    d = R^TK^{-1}[x; 1]
\end{equation}
and normalize it as $\hat d = \frac{d}{\lVert d \rVert}$ resulting in two degrees of freedom. Then, we define the moment $m=C\times \hat d$ encoding the line's offset relative to the origin, giving an additional two degrees of freedom as $m$ has three parameters with the constraint $\hat d \cdot m = 0$. We then let $(\hat d, m)\in\mathbb{R}^6$ be the input to our network together with a one-hot encoded vector of potential classes (only edges and vertex classes here). We tried various alternatives for using Plücker lines in our network. For example, using a separate encoder for the lines and adding it to the decoder in a cross-attention layer between lines and queries, after the cross-attention between points and queries. Another thing we tried was using Rotary Pose Embedding (RoPE) \cite{su2024roformer} in the cross-attention layers between the lines and the queries. The above methods worked as we could predict edges and vertices with acceptable accuracy when we only used lines. However, performance did not improve when lines were combined with points. We are unsure why adding lines did not increase the model accuracy, but one guess is that specifying infinite lines gave too weak a signal. A solution to that problem could have been to use monocular depth to create a finite line segment between the camera center and a 3D point.

\subsection{Fewer One-hot Classes}

Another approach we tested was to limit the size of the input vector to the network. The reasoning behind this was that most of the 177 one-hot classes were rarely used. We tried to assign classes that did not appear in enough pixels or scenes to a dustbin class. However, this did not improve performance. Furthermore, because it was hard to decide on a suitable threshold for when to include a class and when to assign it to the dustbin, we decided not to decrease the number of classes. We believe that the network was likely able to learn to downweight uncommon classes, and thus it did not matter much whether we included them or not. In terms of computational complexity, keeping all classes only increased the computational workload marginally.

\subsection{Vertex Consistency Loss}

Our model predicts edges independently of each other, despite the fact that the wireframe is connected in a graph-like structure. Consequently, predictions corresponding to the same ground truth vertex may still be spatially separated. To address this, we introduced an additional vertex consistency loss.

The Hungarian matching induces correspondences between predicted edge endpoints and ground truth vertices. For each set of endpoint coordinates $\hat{v}_i$ assigned to a specific ground truth vertex $k$, we form the group $G_k = \{ \hat v_i \}$. We compute the mean position $\hat \mu_k = \frac{1}{|G_k|} \sum_{i\in G_k} \hat v_i$ and define a coordinate consistency loss $\mathcal{L}_ {vc}^{k}=\frac{1}{|G_k|}\sum_{i\in G_k}||\hat v_i- \hat \mu_k||_1$. The total loss is the mean over all sets containing at least two edge endpoints. Adding this loss to the model gave roughly a 1\% unit increase on the validation set but did not show any improvement on the public test set, so this was omitted in the final model.

\subsection{Prediction Heads for Post-processing}
In addition to the edge prediction model, two prediction heads were investigated with the goal of simplifying the post-processing stage. The motivation was to let the network directly predict properties of the wireframe instead of relying on heuristics in the post-processing step. In the final model, none of the method provided a significant improvement on the public test set.

\textbf{Vertex feature head.} Our model relies on distance-based merging between vertices that are close enough in space. This can be problematic as wireframes can contain more than one vertex within our distance threshold. To address this, a vertex feature head was added to learn feature embeddings for each predicted vertex. The idea was that vertices belonging to the same ground-truth vertex would get similar feature representations that could make the vertex merging in the post-processing more robust.

The training was performed using a contrastive loss consisting of both positive and negative terms. The positive term encouraged features belonging to the same ground truth vertex to be similar, while the negative term penalized high similarity between features belonging to different vertices. During inference, the vertex features were incorporated into the vertex merging post-processing step by considering both the spatial distance and feature similarity. The rest of the network was frozen during training of this prediction head. Including the vertex features in the post-processing step gave only marginal improvements.

\textbf{Edge class prediction head.} The vertex merging procedure modifies vertex locations to satisfy connectivity constraints between edges. As a side effect, this can change the orientation of the edges and cause initially horizontal edges to lose their horizontal alignment. To mitigate this, the edge classification head was introduced.

The training data contains semantic edge classes, some which correspond to horizontal structures. Based on an analysis of the validation set, the original edge classes were divided into horizontal and non-horizontal categories. The prediction head was trained to predict this label using cross-entropy loss. The edge classification achieved high accuracy, indicating that the distinction between the categories could be learned.

The predicted class was then used during post-processing.  Edges classified as horizontal were constrained to remain horizontal after vertex merging. Despite the high classification accuracy, incorporating this information did not provide any significant improvement on the HSS score.


\section{Conclusion}
\label{sec:conclusion}
In this paper, we present our contribution to the S23DR Challenge 2026, which aims to reconstruct the roof wireframes from a sparse SfM point cloud and semantic image cues. Our approach is based on a DETR \cite{carion2020end} inspired transformer architecture that combines geometric information from the point cloud with semantic features from Gestalt and ADE20k segmentations, fused with a multi-view aggregated local-context Gestalt feature.

To improve the model's performance, we introduce a post-processing stage where we iteratively merge nearby vertices. We further employ test-time augmentation and model ensembling, which provides additional performance gains.

Our method achieves an HSS score of 0.6476 on the private test set, securing second place in the challenge leaderboard and outperforming all but the winning submission.

\vspace{0.5em}
{\small 
\noindent\textbf{Acknowledgments}
The work was supported by ELLIIT, 
the Swedish Research Council (Grant No. 2023-05424), and the Wallenberg AI, Autonomous Systems and
Software Program (WASP) funded by the Knut and Alice Wallenberg Foundation.
Compute was provided by the supercomputing resource Berzelius provided by National Supercomputer Centre at Linköping University and the Knut and Alice Wallenberg foundation.
}

{
    \small
    \bibliographystyle{ieeenat_fullname}
    \bibliography{main}
}


\end{document}